\pdfoutput=1

\documentclass{article}

\usepackage[final,nonatbib]{neurips_2018}

\usepackage[utf8]{inputenc} 
\usepackage[T1]{fontenc}    
\usepackage{hyperref}       
\usepackage{url}            
\usepackage{booktabs}       
\usepackage{amsfonts}       
\usepackage{nicefrac}       
\usepackage{microtype}      
\usepackage{graphicx}
\usepackage[export]{adjustbox}

\title{3D human pose estimation with adaptive receptive fields and dilated temporal convolutions}

\author{
  Eduardo Castillo \\
  Cornell Tech\\
  2 W Loop Rd, New York, NY 10044 \\
  \texttt{ec833@cornell.edu} \\
  \And
  Irene Font Peradejordi \\
  Cornell Tech\\
  2 W Loop Rd, New York, NY 10044 \\
  \texttt{if76@cornell.edu} \\
  \And
  Michael Shin \\
  Cornell Tech\\
  2 W Loop Rd, New York, NY 10044 \\
  \texttt{js3552@cornell.edu} \\
  \And
  Shobhna Jayaraman \\
  Cornell Tech\\
  2 W Loop Rd, New York, NY 10044 \\
  \texttt{sj747@cornell.edu} \\
}

\begin{document}
\maketitle

\begin{abstract}
    In this work, we demonstrate that receptive fields in 3D pose estimation can be effectively specified using optical flow. We introduce adaptive receptive fields, a simple and effective method to aid receptive field selection in pose estimation models based on optical flow inference. We contrast the performance of a benchmark state-of-the-art model running on fixed receptive fields with their adaptive field counterparts. By using a reduced receptive field, our model can process slow-motion sequences (10x longer) 23\% faster than the benchmark model running at regular speed. The reduction in computational cost is achieved while producing a pose prediction accuracy to within 0.36\% of the benchmark model.
\end{abstract}

\section{Introduction}

Three-dimensional pose estimation is a vibrant field of research in deep learning and computer vision. Efficient 3D pose estimation algorithms are extensively used in a variety of areas such as action recognition, virtual reality, and human-computer interaction \cite{pavllo20183d}. Driven by progress in inference accuracy as well as improved image data aggregation and dissemination, these algorithms have gained significant traction in commercial and industrial applications. Noteworthy examples include behavioral inference monitoring in the public safety sector and virtual fitting room implementations in the fashion industry \cite{ribeiro_ferworn_denko_tran_2009, ge2019deepfashion2, 10.3389/fnsys.2019.00020}.

One of the key areas of research within 3D pose estimation focuses on reducing the ambiguity of 2D to 3D mappings in video multimedia. This ambiguity stems from the existence of multiple 3D poses which may be inferred from the same 2D joint keypoints. Previous work tackled this problem by capturing a video’s temporal information with recurrent neural networks \cite{Hossain_2018, 10.1007/978-3-030-01234-2_8}. In our work, we aim to build on the approach of current state-of-the-art models in this space, achieved by Facebook AI Research in their paper “3D human pose estimation in video with temporal convolutions and semi-supervised training” \cite{pavllo20183d}. Facebook AI Research uses a novel approach to solve the ambiguity described above: instead of using a recurrent neural network (RNN), they use a fully convolutional 1D neural network (CNN) that takes 2D joint keypoint sequences as input and generates 3D pose estimates as output. To make sure they capture the long-term video information, they employ dilated convolutions. Their model results in higher accuracy, simplicity, as well as efficiency –both in terms of computational complexity, as well as the number of parameters compared to approaches that rely on RNN model structures \cite{Hossain_2018, 10.1007/978-3-030-01234-2_8}.

Facebook AI Research’s work proposes both a supervised and unsupervised approach using two well-known computer vision datasets: Human3.6M and HumanEva. Collecting labels for 3D human pose estimation is quite resource-intensive as it requires an expensive motion-capture setup as well as lengthy recording sessions. For this reason, their supervised approach is particularly interesting. 

Human3.6M contains 3.6 million video frames for 11 human subjects. Seven of them are annotated with 3D poses. With this data set, \cite{pavllo20183d} manages to outperform the previous best results by 6 mm (an 11\% improvement) in mean per-joint position error. HumanEva-I is a smaller dataset, containing three human subjects recorded from three different camera views. HumanEva-I is also highly cited in the literature. The Human3.6M dataset is recorded at 50 Hz while the HumanEva-I is recorded at 60 Hz. 

Our team performed a deep technical analysis of the temporal dilated convolutional model proposed by Facebook AI Research and introduced a novel element–the adaptive receptive field parameter. We demonstrate that using optical flow to adapt the receptive fields depending on the amount of movement in a video over various sequences can help to reduce computational costs while achieving statistically equivalent mean joint displacement errors.

Results are obtained by contrasting the performance of the state-of-the-art pre-trained model provided by Facebook AI Research on fixed receptive fields with their adaptive receptive field counterparts using Human3.6M videos with modified speeds. Our focus was to compare the baseline model of the video at 1x and 0.5x speeds with regards to the obtained mean per-joint error rate for a subject (subject S5 in our case), over varying receptive fields of 3, 9, 27, 81, and 243 frames.

\section{Related Work}

\subsection{3D Pose Estimation}

Earlier methods for pose estimation revolve around feature extraction, with a focus on immutable factors (such as background scene, lighting, and skin color) from images, and mapping those features to a 3D human pose \cite{1315129, 1634337}. The problem of 3D human pose estimation has been addressed in multiple ways starting from a sequence of 2D human poses. The most successful and efficient approaches for pose estimation follow a consistent routine: (i) Estimate the 2D pose from images, (ii) Map the estimated 2D poses into 3D space.

Many models show that a low-dimensional representation, such as 2D joint keypoints, are powerful enough to estimate 3D poses with high accuracy. Lee and Chen \cite{LEE1985148} were the first to infer 3D poses from their 2D projections given bone length. Through their work, Lee and Chen use a binary decision tree where each branch corresponds to two possible states of a joint relative to its parent. On the other hand, Chen and Ramanan first discussed the idea of a detached 2D pose to search for the nearest neighbor 3D pose within a large database of exemplar poses \cite{chen20163d}.

Moreno-Nouguer \cite{6619310} introduced a novel approach to automatically recover 3D human poses from a single image. They looked to solve the detection of edges, joints, or shadows to infer 3D poses from images. Their solution centered around a Bayesian Framework that integrates a generative model. This generative model was based on latent variables and discriminative 2D part detectors, and 3D inference using a pairwise distance matrix of 2D joints to obtain a distance matrix of 3D joints. And in addition to using multidimensional scaling (MDS) with pose-priors to rule out the ambiguities, this was a consistent attribute which they used to transform ground truth 3D joint positions.

Cheol-hwan et. al \cite{yoo2019fast} improved on Convolutional Neural Networks for 3D hand pose estimation from a single depth image. Since the hand is composed of six different parts, including sequential joints that provide restricted motion, CNNs fall short of modeling the complexity of this structure. To solve this, they propose a Hierarchically Structured Convolutional Recurrent Neural Network (HCRNN) with six branches that estimates the palm and fingers individually. 

When performing frame-by-frame 3D pose estimation, errors independent to each frame can cause jitter. This can be resolved by utilizing temporal information across a sequence of 2D joint positions to estimate a sequence of 3D poses. We discuss temporal dilated convolutional models in the next section of the paper.

\subsection{Temporal dilated convolutional model}

Convolutional models enable parallelization over both the batch and the time dimensions while RNNs cannot be parallelized over time \cite{pavllo20183d}. In CNN models, the path of the gradient between output and input has a fixed length regardless of the sequence length, which mitigates vanishing and exploding gradients which affect RNNs. Moreover, \cite{pavllo20183d} proposes \textit{dilated convolutions} \cite{article} to model long-term dependencies while maintaining computational efficiency.

In an attempt to solve the exploding and vanishing gradients problem and the difficulty of parallelizing the training using an RNN, \cite{cheema2018dilated} already proposes a dilated temporal fully-convolutional neural network (DTFCN) as an automatic framework for semantic segmentation of motion. Additionally, \cite{bai2018empirical} has shown that temporal convolutional networks (TCN) perform just as well, or even better than RNNs in sequencing modeling tasks. Some years before, \cite{article} aalready proved the efficiency of dilated convolutions for semantic segmentation tasks. Their advantage resides in the systematic aggregation of multiscale contextual information without losing resolution. The architecture is based on the fact that dilated convolutions support exponential expansion of the receptive field without loss of resolution or coverage.

To tackle the state saturation problem that LSTMs suffer from, \cite{8461921} propose modeling temporal variations through a stateless dilated convolutional neural network, which uses dilated causal convolution, gated activations, and residual connections. Their work is in the voice-activity detection space where utterance is long, and thus requires the LSTM state to be periodically reset. Their proposed model achieves 14\% improvement in false acceptance rate with a false rejection rate of 1\% over state-of-the-art LSTMs for the voice-activity-detection task.

Other papers successfully use dilated convolutions in tasks like machine translation \cite{kalchbrenner2016neural} and audio generation \cite{oord2016wavenet}. 

\begin{figure}[ht]
    
    \centering
    
    \includegraphics[width=396px]{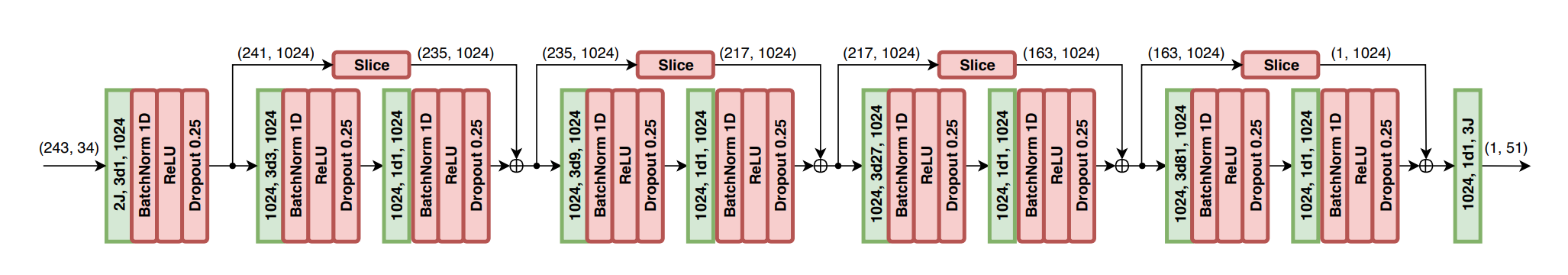}
    
    \caption{An instantiation of the benchmark state-of-the-art model \cite{pavllo20183d}, fully-convolutional 3D pose estimation architecture. The input consists of 2D keypoints for a receptive field of 243 frames (B = 4 blocks) with J = 17 joints. Convolutional layers are in green where 2J, 3d1, 1024 denotes 2 · J input channels, kernels of size 3 with dilation 1, and 1024 output channels. They also show tensor sizes in parentheses for a sample 1-frame prediction, where (243, 34) denotes 243 frames and 34 channels. Due to valid convolutions, they slice the residuals (left and right, symmetrically) to match the shape of subsequent tensors.}
    \label{facebookNN}
\end{figure}

\begin{figure}[ht]
    
    \centering
    
    \includegraphics[width=300px]{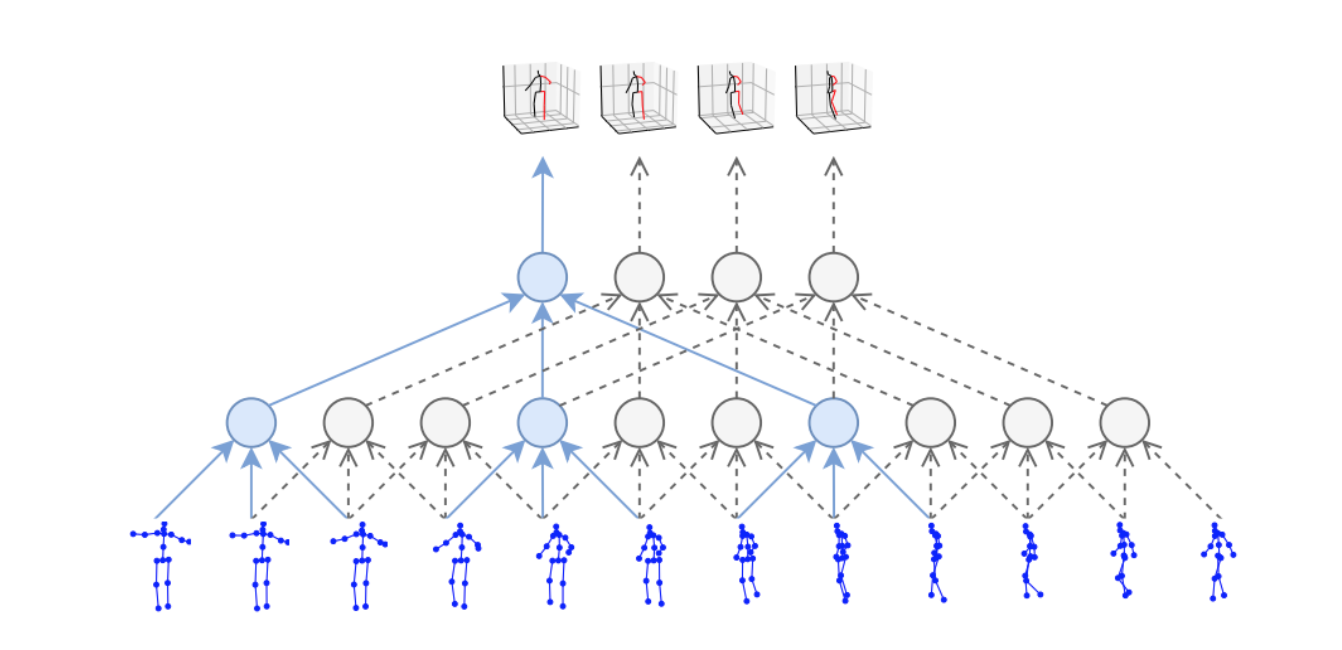}
    
    \caption{The benchmark state-of-the-art model \cite{pavllo20183d} temporal convolutional model takes 2D keypoint sequences (bottom) as input and generates 3D pose estimates as output (top). They employ dilated temporal convolutions to capture long-term information.}
    \label{facebookNNtree}
\end{figure}

The benchmark state-of-the-art model \cite{pavllo20183d} seen in figure \ref{facebookNN} is using an input layer that takes the concatenated (\textit{x,y}) coordinates of the 17 unique joints per frame and applies a temporal convolution with kernel size\textit{ W }and \textit{C} output channels. This is followed by \textit{B} ResNet-style blocks which are surrounded by a skip-connection. Each block performs a 1D convolution with kernel size \textit{W} and dilation factor \[ D= W^2 \] followed by a convolutional with kernel size 1. All convolutional operations, except in the last layer, are followed by batch normalization, ReLU functions, and dropout (p = 0.25). Each block increases the receptive field exponentially by factor \textit{W}, while the number of parameters only increases linearly. See figure \ref{facebookNNtree} for a better understanding of this tree structure. 

\subsection{Optical Flow}

Optical flow estimates the motion of every pixel in a sequence of images. It describes a sparse or dense vector field, where a displacement vector is assigned to a certain pixel position, which points to where that pixel can be found in another image. Mainstream optical flow estimation algorithms can be grouped as follows: region-based matching, differential, and energy-based algorithms.

FlowNet–an optical flow neural network, resolves problems with minor displacements and noisy artifacts in estimated flow fields. In “Evolution of Optical Flow Networks with deep networks” \cite{ilg2016flownet},the authors focus on the training data and discuss elements of scheduling the data presentation and its importance. They developed an architecture that includes warping the second image with an intermediate flow. They introduced a sub-network specializing in small motions to further focus on movement displacements. Both FlowNet1.0 and FlowNet2.0 are end-to-end architectures. FlowNet 2.0 shows decreased estimation error by more than 50\%, but is marginally slower than the FlowNet 1.0. FlowNet 2.0 performed at the same level as standard state-of-the-art methods \cite{ilg2016flownet}.

Jianzhong et al \cite{6182151} propose a method to track the movement of objects. They analyzed many methods which are used to segment Video Objects, and proposed a new algorithm, using optical flow to track objects by using the contours of an object. The Horn–Schunck method of estimating optical flow is a global method that introduces a global constraint of smoothness to solve the aperture problem. The aperture problem states that any varying contours of different orientation moving at varying speeds can cause identical responses in a motion-sensitive neuron in the visual system. In the paper they use this algorithm, to get the position of moving pixels between frames from the velocity vector, in given video streams. Next, they take the contours and extract the object features to calculate the position and velocity values. They achieved accurate, rapid, and stable results with the algorithm to track the moving objects.

\section{Adaptive Receptive Field Implementation - Experimental Setup}

\subsection{Dataset Exploration: Human3.6M Dataset}

For our exploratory research, we heavily relied on the Human3.6M Dataset \cite{6682899} which has 3.6 million 3D human poses and corresponding images, 11 actors (6 Male and 5 Female), and 17 action scenarios (Walking, Eating, Discussion, Phoning). The Human3.6M dataset has a high resolution of 50Hz with 4 different orientations, including accurate 3D Joint Positions and joint angles from a high-speed motion capture system. In addition, it also provides time-of-flight data and laser scans of the actors.

As part of our dataset and pre-processing exploration, we picked Subject S5 as our subject for research. Amongst the actions, this subject pertained to the action “Posing”. As part of our guided methodology in our project, our choice was intended at making a comparative study so that baseline results with the mean per-joint position error for our model could be easily compared and tabulated later (refer to table 1, 2, 3 and 4).

\subsection{Optimized Temporal Convolution Modeling}
The AWS EC2 instance specification we chose are as follows: p2.xlarge, with 4 vCPUs, x86\_64 architecture with a K80 GPU backbone. As our baseline approach for a comparative study, from subject S5’s available 17 actions, we chose the “Posing” action for our research project.

We performed interpolation to get the half-speed samples input to our optimized temporal model. The .cdf in the dataset were converted to .mat (MATLAB) files, then for our 2D and 3D datasets we performed matrix interpolations, ensuring we removed the NaN values for processing the obtained values in the final temporal model.

After loading and converting the Human3.6M 2D frames, we created the 2D poses and saved the joint points in the preparation of the subject S5 dataset with depth, features, and poses attributes.

The model was run for 80 epochs. The architecture of the temporal model exploits temporal information with dilated convolutions over 2D keypoint trajectories from the .npy files.

The default configuration has input features for each joint, while the outputs for each joint are in the dataset files. For the Human 3.6M dataset, the number of output joints is 17. For the filter widths, which determine the number of receptive field frames (i.e. the "number of blocks"), this parameter is input as a flag '--arc 3,3' (for 9 frames) or '--arc 3,3,3' (for 27 frames), etc. during our run of the main script of the model. To note the metrics in the final evaluation step of our research, we measure these given four loss values over time over 80 epochs:

\begin{itemize}
\item E1: Mean per-joint position error (MPJPE) over time which is the mean Euclidean distance between predicted joint positions and ground-truth joint positions.

\item E2: P-Mean per-joint position error which gives the error after alignment with the ground truth in translation, rotation, and scale.

\item E3: N-Mean per joint position error (N-MPJPE) which aligns the predicted poses with the ground-truth only in scale (N-MPJPE) for semi-supervised experiments.

\item Velocity Error which is the mean per-joint velocity error (MPJVE).
\end{itemize}

These error values are reported later as our findings in the result section.

\subsection{Optical Flow Modeling on Humans 3.6 Dataset}

\begin{figure}[ht]
    
    \centering
    
    \includegraphics[width=200px]{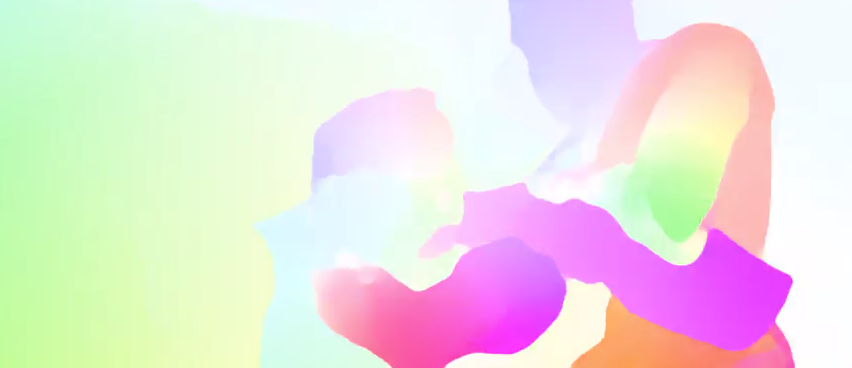}
    
    \includegraphics[width=200px]{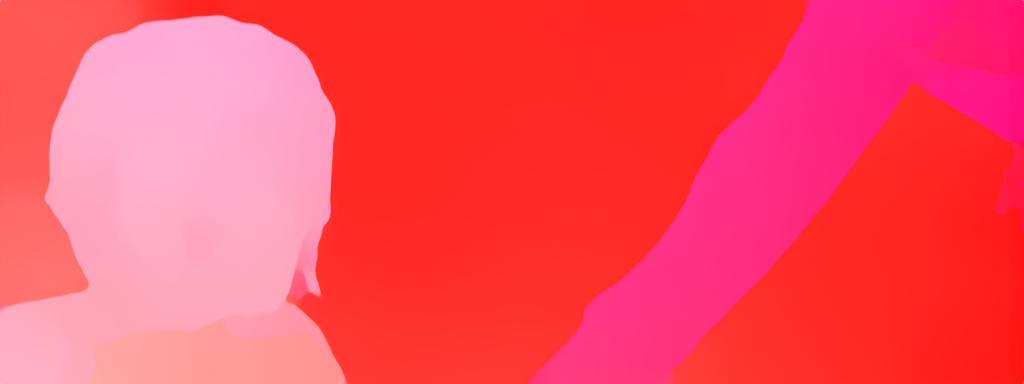}
    \caption{Representative Benchmark Results from FlowNet 2.0 Model}
    \label{sintel}
\end{figure}

We deploy a deep convolutional neural network architecture based on the state-of-the-art FlowNet 2.0 architecture \cite{flownet2-pytorch}. To assure consistency of results with existing literature, models are trained and validated on the reference Sintel benckmark dataset \cite{Butler:ECCV:2012}. As shown in figure \ref{sintel}, detail granularity and resolution consistent with the FlowNet 2.0 results is achieved on the benchmark Sintel dataset.

\subsection{Optical Flow Modeling on Humans 3.6 Dataset}

\begin{figure}[ht]
    
    \centering
    \includegraphics[height=150px, fbox]{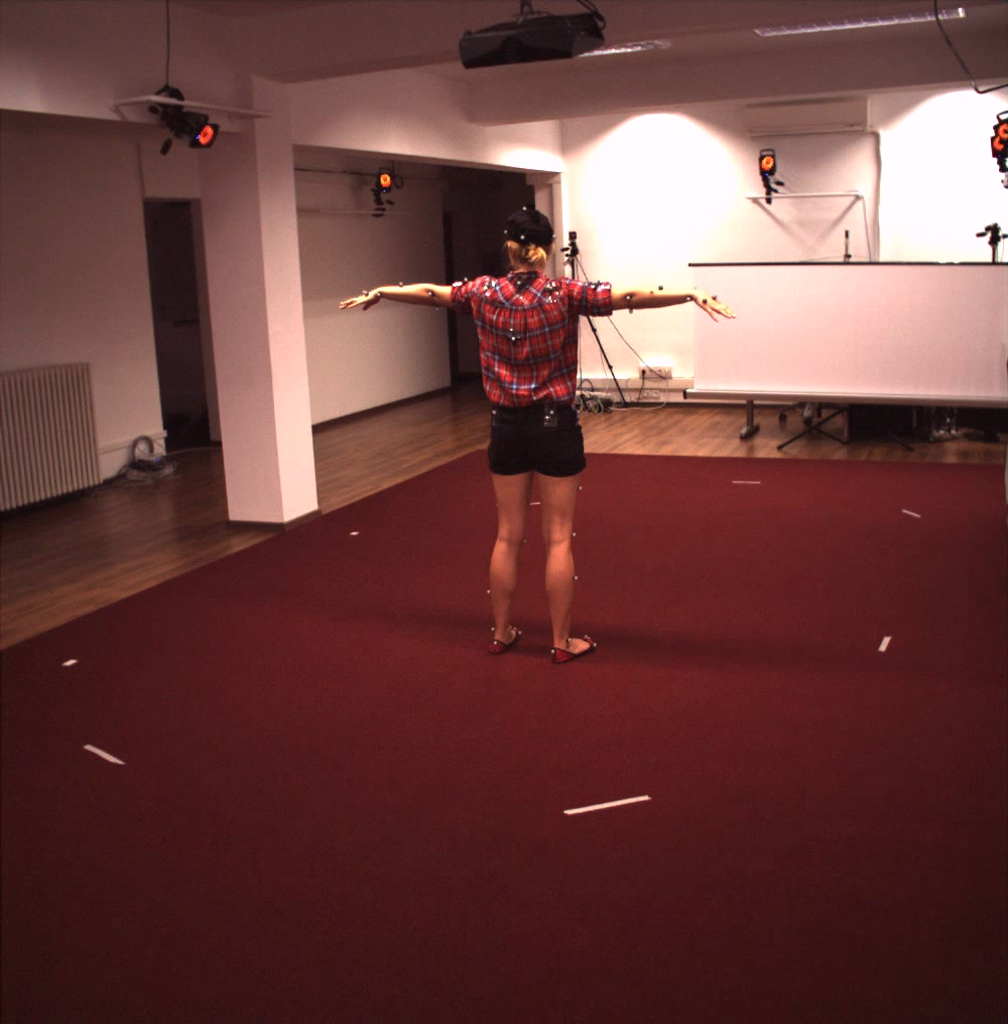}
    \includegraphics[height=150px,fbox]{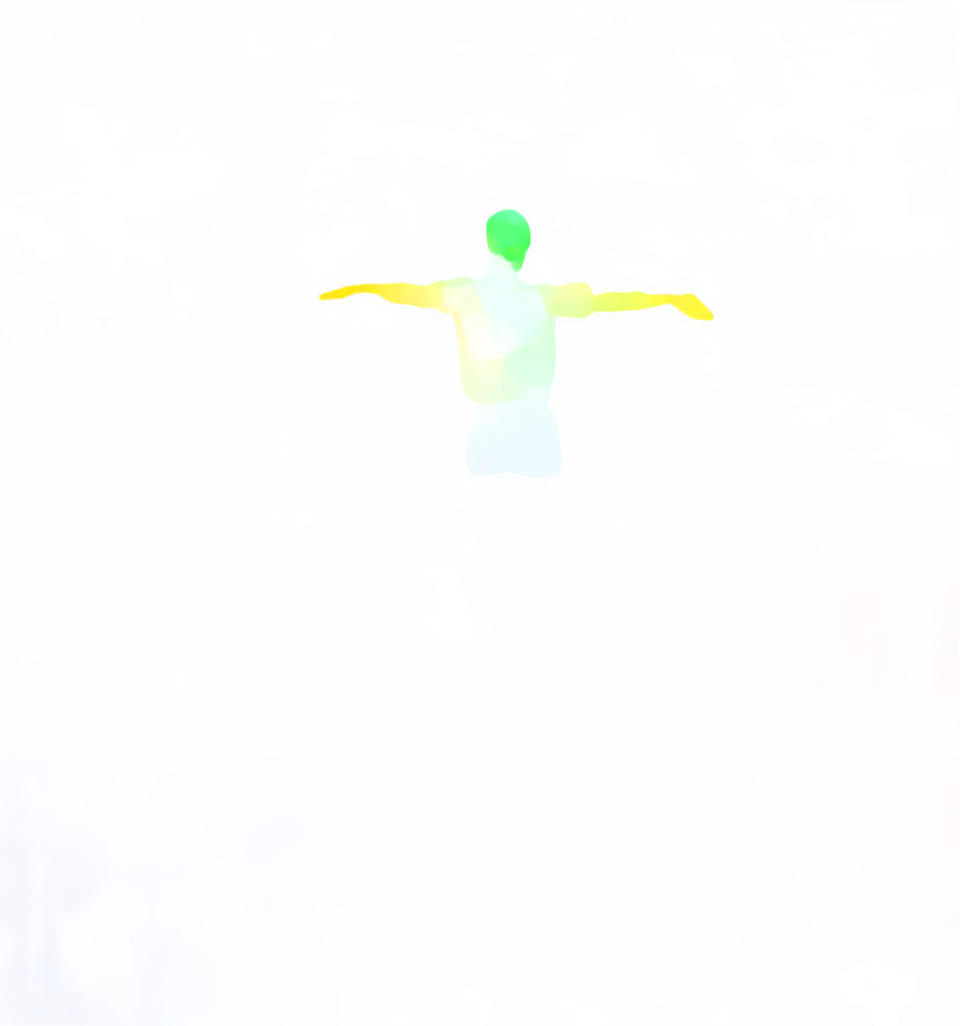}
    \caption{Representative Humans 3.6M Results from FlowNet 2.0 Model}
    \label{h36_1}
\end{figure}

Following benchmarking of our optical flow model, a set of pose estimation videos was selected from our target 3D pose estimation dataset, Humans 3.6M \cite{10.1109/TPAMI.2013.248}. This set of training and testing videos was processed into frames for optical flow inference using our trained FlowNet 2.0 model. Figure \ref{h36_1}, shows representative results obtained through processing of the Humans 3.6M subject "S2" samples.

As shown, detail granularity and resolution on these samples remain adequate and consistent with results on the benchmark dataset. Moreover, high flow regions are indeed localized to areas consistent with subject dynamics, as confirmed by inspection of the raw training videos.

\subsection{Adaptive Receptive Field Regression}

Following model validation on the target dataset, the flow field output of our optical flow model was used to estimate temporal information density across frames–the degree to which motion was present across contiguous pose frames–and produce a best-estimate receptive field parameter for processing of the given sample. 

Several methods of determining the optimal receptive field parameter from the flow field output were tested. First, a single motion value had to be calculated from a frame in the flow field output. The magnitude of the x and y vectors was used to determine a motion value for a single pixel. Then, to integrate these values for each pixel to represent the motion of a frame, various combinations of max-pooling and averaging were attempted. However, the simplest method of taking the max value from the entire frame (clipping outliers) proved to be the most effective. Intuitively, this made sense as most objects move in unison and because the still pixels in a frame shouldn’t affect the motion value. To combine the motion values of all frames to form a motion value for the entire video, we simply averaged the values from each frame.

\section{Results}

The evaluation metrics used are the same as those of the research community, as this made the results comparable. These are the mean per-joint position error (MPJPE) measured in mm. This is the Euclidean distance between the predicted joint position and ground truth joint position.

\begin{figure}[ht]
    
    \centering
    \includegraphics[width=300px]{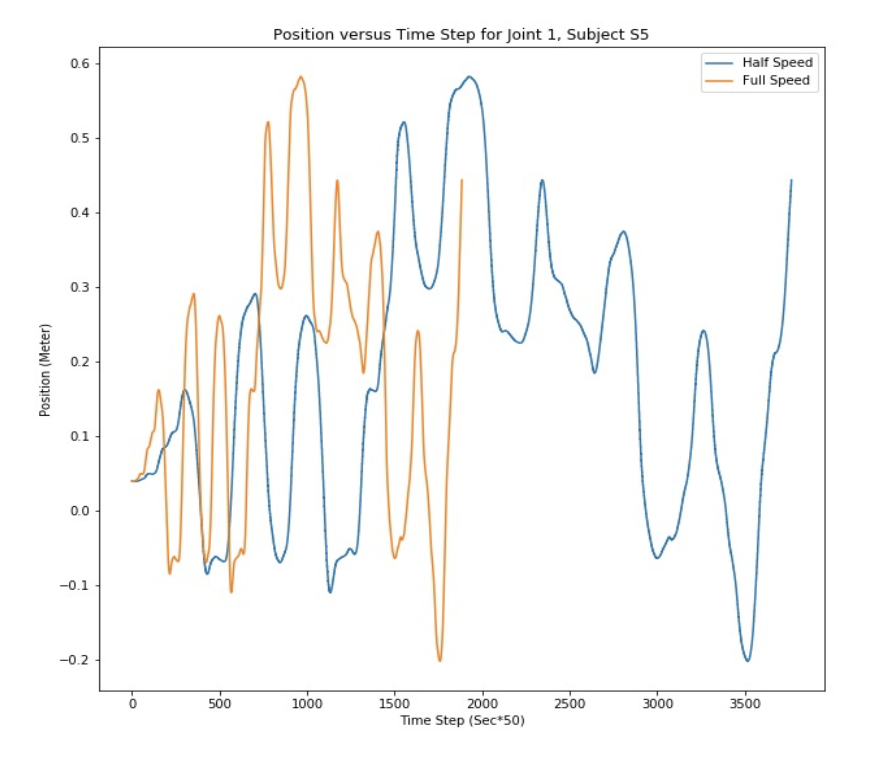}
    \caption{Example of speed reduction video at 0.5x using interpolation}
    \label{slowvideo}
\end{figure}

Experiments were performed for Subject S5 in the “Posing” action at a speed reduction factor of 10x \ref{slowvideo}. Results show that a reduction in receptive field from 243 frames to 81 frames resulted in an MPJPE increase of just 0.36\% or 0.1 mm (28.2 mm to 28.3 mm error). The benchmark state-of-the-art model, which runs on full-speed videos showed an increase in MPJPE of 1.27\% or 0.6 mm (47.1 mm to 47.7 mm error) for the same receptive field parameter change.

For our model, program execution time was reduced by approximately 23\% (56.05 seconds versus 43.4 seconds) by reducing the receptive field as noted above, which further demonstrates the computational advantages of our adaptive model. Furthermore, this result strongly suggests that adaptive receptive fields could offer an effective alternative to fixed receptive field parameterization for systems deployed in highly dynamic environments.

\begin{figure}[ht]
    
    \centering
    \includegraphics[width=300px]{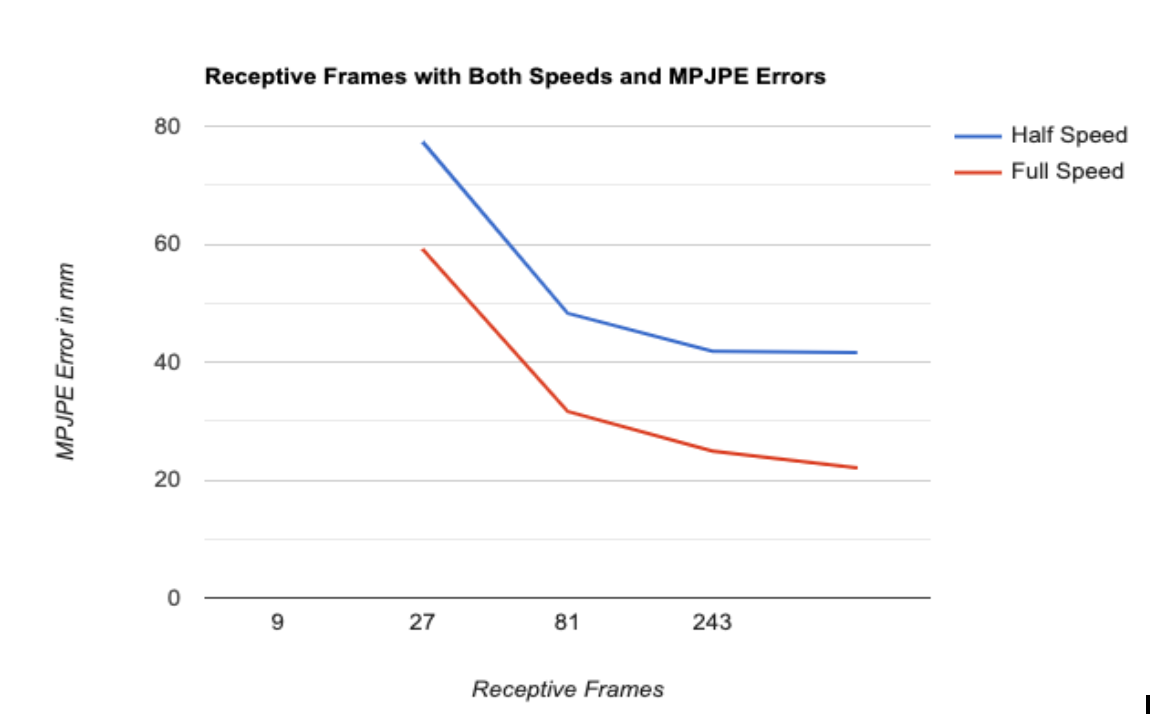}
    \caption{The Figure below depicts Receptive Frames at (9, 27, 81 and 243) vs. the Mean Per-Joint Position True Protocol \# 1 Error at 0.5x speed and 1x Speeds (refer to Table 1)}
    \label{recep}
\end{figure}

\begin{table}
  \caption{True Protocol \# 1 Error (MPJPE) vs. the Receptive Fields in both speeds}
  \label{table1}
  \centering
  \begin{tabular}{lll}
    \toprule
    \cmidrule(r){1-2}
    RECEPTIVE FIELDS & HALF SPEED (mm)     & FULL SPEED (mm) \\
    \midrule
    9 frames   & 77.369 & 59.146 \\
    27 frames  & 48.267 & 31.609 \\
    81 frames  & 41.845 & 24.866 \\
    243 frames & 41.608 & 22.077 \\
    \bottomrule
  \end{tabular}
\end{table}

\begin{table}
  \caption{True Protocol \# 2 Error (P-MPJPE) vs. the Receptive Fields in both speeds}
  \label{table2}
  \centering
  \begin{tabular}{lll}
    \toprule
    \cmidrule(r){1-2}
    RECEPTIVE FIELDS     & HALF SPEED (mm)     & FULL SPEED (mm) \\
    \midrule
    9 frames   & 44.057 & 42.219 \\
    27 frames  & 33.528 & 25.986 \\
    81 frames  & 31.202 & 20.110 \\
    243 frames & 31.160 & 17.567 \\
    \bottomrule
  \end{tabular}
\end{table}

\begin{table}
  \caption{True Protocol \# 3 Error (N-MPJPE) vs. the Receptive Fields in both speeds}
  \label{table3}
  \centering
  \begin{tabular}{lll}
    \toprule
    \cmidrule(r){1-2}
    RECEPTIVE FIELDS     & HALF SPEED (mm)     & FULL SPEED (mm) \\
    \midrule
    9 frames   & 60.420 & 50.651 \\
    27 frames  & 42.422 & 30.785 \\
    81 frames  & 39.078 & 24.667 \\
    243 frames & 39.764 & 21.976 \\
    \bottomrule
  \end{tabular}
\end{table}

\begin{table}
  \caption{Velocity (MPJVE) ERROR in both speeds}
  \label{table4}
  \centering
  \begin{tabular}{lll}
    \toprule
    \cmidrule(r){1-2}
    RECEPTIVE FIELDS     & HALF SPEED (mm)     & FULL SPEED (mm) \\
    \midrule
    9 frames   & 2.713 & 2.308 \\
    27 frames  & 2.600 & 2.094 \\
    81 frames  & 2.473 & 1.861 \\
    243 frames & 2.444 & 1.749 \\
    \bottomrule
  \end{tabular}
\end{table}

\clearpage

\section{Conclusions}

In this work, we demonstrate that receptive fields in 3D pose estimation can be effectively specified using optical flow, which estimates the motion of every pixel in a sequence of images. By doing so, computational costs can be effectively decreased in low movement sequences while maintaining equivalent performance. 

Experiments performed using lower speed videos–modified using keypoints interpolation–of the Subject S5 (action “Posing”) of the Human3.6M dataset successfully shows that a reduction in receptive field from 243 frames to 81 frames resulted in an MPJPE increase of just 0.36\% or 0.1 mm (28.2 mm to 28.3 mm error). The benchmark state-of-the-art model from Facebook AI research \cite{pavllo20183d}, which runs on full-speed videos showed an increase in MPJPE of 1.27\% or 0.6 mm (47.1 mm to 47.7 mm error) for the same receptive field parameter change.

Our proposed model execution time was lowered by approximately 23\% (56.05 seconds versus 43.4 seconds) by reducing the receptive field as noted above, which further demonstrates the computational advantages of our adaptive model. Furthermore, this result strongly suggests that adaptive receptive fields could offer an effective alternative to fixed receptive field parameterization for systems deployed in highly dynamic environments.

\section{About the Team Members}

\textbf{Michael Shin, CS’20}
Michael has a bachelor’s degree in Computer Engineering from The University of Michigan. Prior to joining Cornell Tech, he worked as a GPU engineer for Intel creating software simulators of various pixel pipeline components for validation purposes. Since then he’s reinvented himself as an entrepreneur and developer specializing in the UX and development of iOS and Android apps. Michael is originally from Sydney, Australia.

\textbf{Eduardo Castillo, ECE’20}
Eduardo has a bachelor’s degree in Mechanical Engineering, summa cum laude from Rowan University. Prior to joining 	Cornell Tech, he worked as a full-time design engineer at PSEG Nuclear developing design upgrades for large power generation assets. He is also an admitted fellow at the MIT System Design and Management program. Eduardo continues to work for PSEG Nuclear on a part-time basis and is also a small business owner in the cosmetics space. Eduardo is originally from Santo Domingo, Dominican Republic.

\textbf{Irene Font Peradejordi, CM’21}
Irene has a bachelor’s degree in Design \& Advertising from Pompeu Fabra University in Barcelona. Prior to joining Cornell Tech, she earned a MSc in Cognitive Sciences and Artificial Intelligence from Tilburg University in The Netherlands. She worked in Barcelona and Boston in a MIT Media Lab startup and she helped found an AI community (Saturdays.AI) that expanded to more than 25 cities in Spain and Latin America. She is part of “la Caixa” Merit fully-funded Fellowship program.

\textbf{Shobhna Jayaraman, CS’20}
Shobhna has a bachelor’s degree in Computer Science and Engineering from Amity University. Prior to joining Cornell Tech, she worked full-time as a software engineer/ DevOps engineer for Orange Business Services, Gurgaon, India. She has interned as a Data Analyst with several companies. During her bachelor’s, she wrote two machine learning focussed research papers which have been published in IEEE Xplore Journal.

\subsubsection*{Acknowledgments}

We would like to express our deep gratitude to Dr. Jin Sun and Dr. Christopher Kanan, for their patient guidance, enthusiastic encouragement and insightful feedback during this project. Their engaging instruction, dynamic leadership and commitment to our success have been a beacon of hope for all of us during these unprecedented times.

The project that gave rise to these results was partially supported by a fellowship from ”la Caixa” Foundation (ID 100010434) for Post-Graduate Studies. The fellowship code is LCF/BQ/AA18/11680107 (Irene Font Peradejordi).

\bibliography{bib}{}

\begin{thebibliography}{10}

\bibitem{pavllo20183d}
Dario Pavllo, Christoph Feichtenhofer, David Grangier, and Michael Auli.
\newblock 3d human pose estimation in video with temporal convolutions and
  semi-supervised training, 2018.

\bibitem{ribeiro_ferworn_denko_tran_2009}
Cristina Ribeiro, Alexander Ferworn, Mieso Denko, and James Tran.
\newblock Canine pose estimation: A computing for public safety solution.
\newblock {\em 2009 Canadian Conference on Computer and Robot Vision}, 2009.

\bibitem{ge2019deepfashion2}
Yuying Ge, Ruimao Zhang, Lingyun Wu, Xiaogang Wang, Xiaoou Tang, and Ping Luo.
\newblock Deepfashion2: A versatile benchmark for detection, pose estimation,
  segmentation and re-identification of clothing images, 2019.

\bibitem{10.3389/fnsys.2019.00020}
Ahmet Arac, Pingping Zhao, Bruce~H. Dobkin, S.~Thomas Carmichael, and Peyman
  Golshani.
\newblock Deepbehavior: A deep learning toolbox for automated analysis of
  animal and human behavior imaging data.
\newblock {\em Frontiers in Systems Neuroscience}, 13, 2019.

\bibitem{Hossain_2018}
Mir Rayat~Imtiaz Hossain and James~J. Little.
\newblock Exploiting temporal information for 3d human pose estimation.
\newblock {\em Lecture Notes in Computer Science}, page 69–86, 2018.

\bibitem{10.1007/978-3-030-01234-2_8}
Kyoungoh Lee, Inwoong Lee, and Sanghoon Lee.
\newblock Propagating lstm: 3d pose estimation based on joint interdependency.
\newblock In Vittorio Ferrari, Martial Hebert, Cristian Sminchisescu, and Yair
  Weiss, editors, {\em Computer Vision -- ECCV 2018}, pages 123--141, Cham,
  2018. Springer International Publishing.

\bibitem{1315129}
Proceedings of the 2004 ieee computer society conference on computer vision and
  pattern recognition.
\newblock In {\em Proceedings of the 2004 IEEE Computer Society Conference on
  Computer Vision and Pattern Recognition, 2004. CVPR 2004.}, volume~2, 2004.

\bibitem{1634337}
G.~{Mori} and J.~{Malik}.
\newblock Recovering 3d human body configurations using shape contexts.
\newblock {\em IEEE Transactions on Pattern Analysis and Machine Intelligence},
  28(7):1052--1062, 2006.

\bibitem{LEE1985148}
Hsi-Jian Lee and Zen Chen.
\newblock Determination of 3d human body postures from a single view.
\newblock {\em Computer Vision, Graphics, and Image Processing}, 30(2):148 --
  168, 1985.

\bibitem{chen20163d}
Ching-Hang Chen and Deva Ramanan.
\newblock 3d human pose estimation = 2d pose estimation + matching, 2016.

\bibitem{6619310}
E.~{Simo-Serra}, A.~{Quattoni}, C.~{Torras}, and F.~{Moreno-Noguer}.
\newblock A joint model for 2d and 3d pose estimation from a single image.
\newblock In {\em 2013 IEEE Conference on Computer Vision and Pattern
  Recognition}, pages 3634--3641, 2013.

\bibitem{yoo2019fast}
Cheol hwan Yoo, Seo won Ji, Yong goo Shin, Seung wook Kim, and Sung jea Ko.
\newblock Fast and accurate 3d hand pose estimation via recurrent neural
  network for capturing hand articulations, 2019.

\bibitem{article}
Matthias Holschneider, Richard Kronland-Martinet, J.~Morlet, and
  Ph~Tchamitchian.
\newblock A real-time algorithm for signal analysis with the help of the
  wavelet transform.
\newblock {\em Wavelets, Time-Frequency Methods and Phase Space}, -1:286, 01
  1989.

\bibitem{cheema2018dilated}
Noshaba Cheema, Somayeh Hosseini, Janis Sprenger, Erik Herrmann, Han Du, Klaus
  Fischer, and Philipp Slusallek.
\newblock Dilated temporal fully-convolutional network for semantic
  segmentation of motion capture data, 2018.

\bibitem{bai2018empirical}
Shaojie Bai, J.~Zico Kolter, and Vladlen Koltun.
\newblock An empirical evaluation of generic convolutional and recurrent
  networks for sequence modeling, 2018.

\bibitem{8461921}
S.~{Chang}, B.~{Li}, G.~{Simko}, T.~N. {Sainath}, A.~{Tripathi}, A.~{van den
  Oord}, and O.~{Vinyals}.
\newblock Temporal modeling using dilated convolution and gating for
  voice-activity-detection.
\newblock In {\em 2018 IEEE International Conference on Acoustics, Speech and
  Signal Processing (ICASSP)}, pages 5549--5553, 2018.

\bibitem{kalchbrenner2016neural}
Nal Kalchbrenner, Lasse Espeholt, Karen Simonyan, Aaron van~den Oord, Alex
  Graves, and Koray Kavukcuoglu.
\newblock Neural machine translation in linear time, 2016.

\bibitem{oord2016wavenet}
Aaron van~den Oord, Sander Dieleman, Heiga Zen, Karen Simonyan, Oriol Vinyals,
  Alex Graves, Nal Kalchbrenner, Andrew Senior, and Koray Kavukcuoglu.
\newblock Wavenet: A generative model for raw audio, 2016.

\bibitem{ilg2016flownet}
Eddy Ilg, Nikolaus Mayer, Tonmoy Saikia, Margret Keuper, Alexey Dosovitskiy,
  and Thomas Brox.
\newblock Flownet 2.0: Evolution of optical flow estimation with deep networks,
  2016.

\bibitem{6182151}
{Zhiwen Chen}, {Jianzhong Cao}, {Yao Tang}, and {Linao Tang}.
\newblock Tracking of moving object based on optical flow detection.
\newblock In {\em Proceedings of 2011 International Conference on Computer
  Science and Network Technology}, volume~2, pages 1096--1099, 2011.

\bibitem{6682899}
C.~{Ionescu}, D.~{Papava}, V.~{Olaru}, and C.~{Sminchisescu}.
\newblock Human3.6m: Large scale datasets and predictive methods for 3d human
  sensing in natural environments.
\newblock {\em IEEE Transactions on Pattern Analysis and Machine Intelligence},
  36(7):1325--1339, 2014.

\bibitem{flownet2-pytorch}
Fitsum Reda, Robert Pottorff, Jon Barker, and Bryan Catanzaro.
\newblock flownet2-pytorch: Pytorch implementation of flownet 2.0: Evolution of
  optical flow estimation with deep networks.
\newblock \url{https://github.com/NVIDIA/flownet2-pytorch}, 2017.

\bibitem{Butler:ECCV:2012}
D.~J. Butler, J.~Wulff, G.~B. Stanley, and M.~J. Black.
\newblock A naturalistic open source movie for optical flow evaluation.
\newblock In {A. Fitzgibbon et al. (Eds.)}, editor, {\em European Conf. on
  Computer Vision (ECCV)}, Part IV, LNCS 7577, pages 611--625. Springer-Verlag,
  October 2012.

\bibitem{10.1109/TPAMI.2013.248}
Catalin Ionescu, Dragos Papava, Vlad Olaru, and Cristian Sminchisescu.
\newblock Human3.6m: Large scale datasets and predictive methods for 3d human
  sensing in natural environments.
\newblock {\em IEEE Trans. Pattern Anal. Mach. Intell.}, 36(7):1325–1339,
  July 2014.

\end{thebibliography}
\bibliographystyle{unsrt}

\end{document}